\documentclass[letterpaper, 10 pt, conference]{ieeeconf}  
\IEEEoverridecommandlockouts                             

\usepackage{amsfonts}
\usepackage{amsmath}
\usepackage{amssymb}
\usepackage[english]{babel}
\usepackage{booktabs}
\usepackage[colorlinks=true, citecolor=black, linkcolor=black, urlcolor=black]{hyperref}
\usepackage[capitalize,noabbrev]{cleveref}
\usepackage{comment}
\usepackage{lipsum} 
\usepackage{mathtools}
\usepackage{multirow}
\usepackage{nicefrac}
\usepackage{setspace}

\usepackage[
    detect-all=true,
    detect-family=true,
    group-digits=false,
    separate-uncertainty=true,
    retain-zero-uncertainty=true
]{siunitx}
\usepackage{tabu}
\usepackage{tabularx}
\usepackage{url}

\usepackage{vector} 

\usepackage[
    style=ieee,
    backend=biber,
    url=false,
    doi=false,
    natbib=true,
    mincitenames=1,
    dashed=false,
    isbn=false
]{biblatex}
\usepackage[capitalize]{cleveref}

\AtEveryBibitem{\clearfield{issn}}
\AtEveryCitekey{\clearfield{issn}}

\usepackage{color}
\usepackage{graphicx}
\usepackage{pgfplots}
\pgfplotsset{compat=newest}
\pgfplotsset{every axis legend/.append style={legend cell align=left}}
\usepackage[mode=buildnew]{standalone}
\usepackage{subcaption}
\usepackage{tikz}
\usetikzlibrary{shapes, arrows, arrows.meta, external, positioning}
\usepackage{wrapfig}

\definecolor{vandeusen}{RGB}{73,92,111}
\definecolor{cordovan}{RGB}{152,68,71}
\definecolor{pastelBlue}{RGB}{0,114,178}
\definecolor{pastelRed}{RGB}{245,97,92}
\definecolor{pastelGreen}{RGB}{0,158,115}
\definecolor{pastelPurple}{RGB}{135,112,254}

\renewrobustcmd{\bfseries}{\fontseries{b}\selectfont}
\renewrobustcmd{\boldmath}{}
\newrobustcmd{\B}{\bfseries}

\newcommand{\miniboone}{\textsc{Miniboone}}
\newcommand{\hepmass}{\textsc{Hepmass}}
\newcommand{\bsds}{\textsc{Bsds300}}

\newcommand{\fashion}{\textsc{Fashion}}
\newcommand{\cifar}{\textsc{Cifar-10}}
\newcommand{\celeba}{\textsc{Celeba}}

\DeclareTextFontCommand{\textsc}{\fontfamily{cmr}\fontseries{m}\selectfont\scshape}
\usepgfplotslibrary{external}
\IEEEoverridecommandlockouts                              

\title{\LARGE \bf
Simplifying Flow Matching Transformations with \\ Low-Rank Mixture Models}

\author{Liam A. Kruse, Houjun Liu, Alexandros E. Tzikas, Mansur M. Arief, and Mykel J. Kochenderfer%
\thanks{L. A. Kruse (corresponding author), H. Liu, A. E. Tzikas, and M. J. Kochenderfer are with the Stanford Intelligent Systems Laboratory in the Department of Aeronautics and Astronautics at Stanford University, Stanford, CA 94305, USA (email: \{lkruse, houjun, alextzik, mykel\}@stanford.edu).}
\thanks{M. Arief is with the Industrial and Systems Engineering Department at King Fahd University of Petroleum and Minerals (KFUPM), Dhahran, 31261, KSA (email: mansur.arief@kfupm.edu.sa).}
}

\begin{document}

\maketitle
\thispagestyle{empty}
\pagestyle{empty}

\begin{abstract}%
\label{abstract}
\looseness=-1 Normalizing flows are powerful generative models that learn an invertible mapping between complex data distributions and simple latent distributions, typically a standard normal density. 
However, this choice of latent density can impose unnecessary complexity on the learned flow transformation due to the topological mismatch between the latent and data densities, leading to slower training and suboptimal performance. 
In this work, we propose using mixtures of probabilistic principal component analyzers (MPPCA) as the latent density for normalizing flows. 
We simplify the learned flow transformation by learning a latent distribution that more closely aligns with the data distribution in terms of KL divergence, thus enabling faster convergence and improved generative performance.
Critically, MPPCA models can be fit quickly and cheaply using the expectation-maximization algorithm, making them a practical choice for initializing latent distributions even in high-dimensional generative tasks.
We validate our method on both tabular and image datasets, demonstrating consistent gains in training efficiency and generation quality compared to baselines.
\end{abstract}

\section{Introduction}
\label{introduction}
Normalizing flows are a class of generative models that learn a deterministic, invertible mapping between a complex data density and a simpler \textit{base} density, enabling both sampling and density estimation \citep{rezende2015variational}.
\textit{Continuous} normalizing flows (CNFs) extend the framework by parameterizing the transformation as the solution to an ordinary differential equation (ODE) \citep{chen2018neural}.
The learned vector field, e.g., the \textit{integration map}, induces a pushforward probability distribution as it transports samples from the base density to the target data density.
Early CNF models were trained via maximum likelihood estimation, requiring expensive simulation through the learned ODE dynamics \citep{grathwohl2018ffjord}.
Recently, the development of the \textit{flow matching objective} allowed flows to be trained in a simulation-free approach by directly regressing to the ODE vector field, akin to the approach used to train diffusion models \citep{lipmanflow}.
Flow matching (FM) has enabled scalable flows to perform high-dimensional tasks such as image generation, image translation, and learning single-cell dynamics \citep{tongimproving}.

\begin{figure*}
    \centering
    \includegraphics[width=0.99\textwidth]{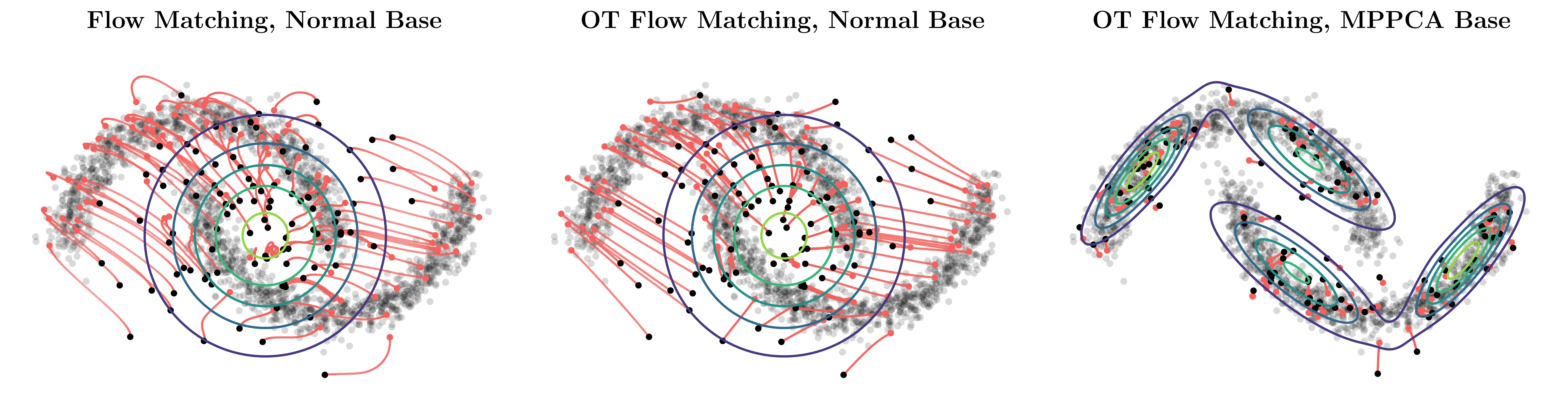}
    \caption{The integration paths through the flow model are not unique and might be unnecessarily complex (left). Flow matching with OT principles produces straighter paths (center). We propose low-rank base densities to further simplify the learned flow transformation (right).}
    \label{fig:fig1}
\end{figure*}

Unfortunately, the learned integration map is not unique, as different dynamics and integration paths can still induce the same pushforward density. 
Transformations with unnecessarily complex dynamics can increase the computational cost at inference time.
Poorly conditioned dynamics require more time steps for an adaptive-step solver to solve the ODE \citep{finlay2020train} and result in larger time-discretization errors for fixed-step solvers, ultimately degrading the quality of the generated samples \citep{liuflow}.
To address this shortcoming of CNF training, inductive biases have been introduced to encourage simpler and more efficient transformations.
For example, optimal transport (OT) principles have been incorporated to regularize the learned dynamics \citep{finlay2020train} and to encourage straighter paths between the data and base densities \cite{tongimproving, pooladian2023multisample}.

In this work, we learn expressive base distributions to simplify the learned flow transformation and provide an inductive bias on the training procedure, as shown in \cref{fig:fig1}.
While flow models typically rely on simple standard normal base densities to permit tractable density estimation, we propose mixtures of probabilistic principal component analyzers (MPPCA) \citep{tipping1999mixtures}---a type of low-rank Gaussian mixture model---as a more flexible alternative.
MPPCA models offer two key characteristics: 1) they have an analytical likelihood expression, preserving the computational efficiency required for density estimation, and 2) they can be fit efficiently even on high-dimensional data using the expectation-maximization (EM) algorithm. 
By initializing the base distribution to closely approximate the target data distribution, our approach reduces the complexity of the learned CNF transformation, leading to faster convergence and improved generative performance.
In practice, our approach \textit{warm-starts} the flow training procedure, trading many iterations of gradient-based optimization for a small number of efficient EM steps.
Our specific contributions include the following:
\begin{itemize}
    \item We construct expressive base distributions for flows using MPPCA models and demonstrate that the MPPCA fitting procedure is computationally inexpensive compared to the flow training procedure.
    \item \looseness=-1 We show that MPPCA base distributions can be combined with optimal transport-based flow matching objectives to generate higher-quality samples while reducing inference-time integration steps.
    \item We quantitatively evaluate our approach on challenging density estimation and generative tasks, including training on high-dimensional image datasets.
\end{itemize}

\section{Related Work}
\label{sect:related-work}
Several prior works have investigated the role of the base density on flow performance and training characteristics. 
\citet{papamakarios2017masked} compared standard normal base densities against a Masked Autoencoder for Distribution Estimation \citep{germain2015made}, finding that the standard normal density was competitive on most datasets. 
Researchers have noted that Gaussian mixture model (GMM) base densities can improve classification performance, as data samples from a particular class are mapped to a specific latent mixture component \cite{ardizzone2020training, izmailov2020semi}. 
In this work we are not concerned with classification, but rather unconditional density estimation and sampling.
Furthermore, \citet{ardizzone2020training} assume that the mixture components have unit covariance matrices,  while \citet{izmailov2020semi} initialize the mixture means randomly and do not fit them along with the flow parameters. 
We learn base densities that are close to the true data density, and fit the MPPCA models using the EM algorithm prior to training the flow. 
\citet{hagemann2021stabilizing} also use GMMs as base densities, and demonstrate that splitting the base density into multiple modes can control the Lipschitz constants of invertible neural networks. 
They use heuristics to choose the GMM component means, whereas we fit ours directly to the data density. 
\citet{stimper2022resampling} develop a base distribution for normalizing flows based on learned rejection sampling. Their objective is to better model target distributions with complex topological structure supports. However, their base distribution cannot be directly evaluated since the normalization constant is unknown. In contrast, MPPCA models admit exact likelihood calculations, preserving the density estimation capabilities of the entire flow model.

Intuitively, an expressive base density serves as an inductive bias that simplifes the learned flow transformation by reducing the learning requirements of the flow. 
Researchers have investigated other strategies to simplify the flow transformation, notably focusing on the connections between optimal transport and the integration paths. 
Regularizing CNFs with a transport cost can straighten the integration paths and reduce the computational cost of simulation-based maximum likelihood training procedures \cite{finlay2020train,onken2021ot}.
More recently, research groups have exploited minibatch approximations of the OT map between two distributions to produce straighter flows, enhancing both training and inference performance \cite{pooladian2023multisample,tongimproving}.
Meanwhile, Liu et al. \cite{liuflow} solve a nonlinear least squares optimization problem to find \textit{straight} couplings---rather than \textit{optimal} couplings---which can be integrated in a single Euler step.

Learning expressive mixture models in high dimensions is a challenging task.
High-dimensional covariance estimates risk overfitting to noise, and the matrices can become ill-conditioned or singular \cite{ledoit2004well}.
Furthermore, storing full-rank matrices requires memory that scales quadratically with the number of dimensions.
One solution is to constrain the learned covariance matrices to be low-rank, effectively modeling the covariances as full-rank matrices on a learned low-dimensional subspace \cite{tipping1999mixtures}.
Two common frameworks for modeling low-rank GMMs include mixtures of probabilistic component analyzers \cite{tipping1999mixtures} and mixtures of factor analyzers (MFAs) \cite{ghahramani1996algorithm}.
In this work, we focus on MPPCA models since they can be fit with closed-form expectation-maximization updates.
\citet{richardson2018gans} use MFA models for image generation, demonstrating that low-rank mixture models can be trained on full-sized images despite the high dimensionality.
Covariance matrix adaptation (CMA) is an evolutionary optimization algorithm that iteratively adapts a covariance matrix to improve sample efficiency \cite{hansen2016cma}. 
Low-rank updates make the optimization process robust and sample efficient \cite{kochenderfer2019algorithms}.

\section{Continuous Normalizing Flows}
\label{sect:flows}
We outline the fundamental theory behind continuous normalizing flows, the flow matching objective, and flow-based generative modeling.

\subsection{Continuous Normalizing Flows}
\label{sect:cnfs}
A smooth, time-varying vector field $u_t : [0, 1] \times \mathbb{R}^d \rightarrow \mathbb{R}^d$ defines an ordinary differential equation (ODE):
\begin{equation}
dx = u_t(x) dt
\end{equation}
We denote the ODE solution as $\phi_t (x)$, which is a diffeomorphic map that transports a point $x$ along the vector field from time $0$ up to time $t$ and has initial conditions $\phi_0(x) = x$.
Given a probability distribution $p_0(x)$, the map $\phi_t (x)$ induces a \textit{pushforward} density
\begin{equation}
\label{eq:pushforward}
p_t(x) = p_0(\phi^{-1}_t (x)) \left\lvert \det \left[ \dfrac{\partial \phi^{-1}_t}{\partial x} (x)\right]\right\rvert
\end{equation}
This time-varying density is characterized by the \textit{continuity equation} \citep{tongimproving}
\begin{equation}
\label{eq:continuity}
\dfrac{\partial p}{\partial t} = - \nabla \cdot (p_t u_t)
\end{equation}
\citet{chen2018neural} modeled the vector field $u_t$ with a neural network, creating a deep parametric generative model called a continuous normalizing flow (CNF). 
Samples from $p_t(x)$ can be obtained by drawing samples $x \sim p_0(x)$ and transporting them along the vector field $u$.
Furthermore, the density $p_t(x)$ can be evaluated using the change-of-variables formula from \cref{eq:pushforward}.
Previously, normalizing flow generative models were created by stacking discrete sets of diffeomorphic transformations \cite{rezende2015variational}.
Often, the transformations had heavily restricted architectures to ensure invertibility or tractable Jacobian calculations \cite{papamakarios2021normalizing}.
However, CNFs permit unrestricted neural network architectures \cite{grathwohl2018ffjord}.

\subsection{Flow Matching}
Early CNF models were trained by maximum likelihood---a computationally expensive procedure that requires simulating the ODE dynamics at every training step to obtain the instantaneous changes in log density \cite{grathwohl2018ffjord}.
The \textit{flow matching objective} offers a simulation-free approach to training CNFs by regressing to $u$ directly, in a similar fashion to regressing the drift for stochastic differential equations in diffusion models \cite{albergo2023stochastic}.
Consider a mixture of probability paths $p_t\left(x \mid z\right)$ that vary according to a conditional variable $z$.
We can rewrite the marginal probability path $p_t(x)$ as
\begin{equation}
p_t(x) = \int p_t\left(x \mid z\right)  q(z) dz
\end{equation}
where $q(z)$ is a distribution over the conditioning variable.
If the vector field $u_t\left(x \mid z\right)$ generates the path $p_t\left(x \mid z\right)$ from initial conditions $p_0\left(x \mid z\right)$, then the vector field 
\begin{equation}
\label{eq:vectorfield}
u_t(x) = \mathbb{E}_{q(z)} \dfrac{u_t\left(x \mid z \right) p_t\left(x \mid z\right)}{p_t(x)}
\end{equation}
generates the path $p_t(x)$, provided that $p_t$ and $u_t$ satisfy \cref{eq:continuity}.
When the conditional probability paths $p_t\left(x \mid z\right)$ and vector fields $u_t\left(x \mid z \right)$ are known and have simple forms, we can obtain an unbiased stochastic objective for regressing the vector field in \cref{eq:vectorfield} \cite{neklyudov2023action,lipmanflow,albergo2023stochastic,tongimproving}.
Let $v_\theta(t, x) : [0, 1] \times \mathbb{R}^d \rightarrow \mathbb{R}^d$ denote a time-varying vector field parameterized by a neural network with weights $\theta$.
The \textit{conditional} flow matching objective
\begin{equation}
\mathcal{L}_{\mathrm{CFM}}(\theta) = \mathbb{E}_{t,~q(z),~p_t\left(x \mid z\right)} \lVert v_\theta(t, x) - u_t(x \mid z) \rVert^2
\end{equation}
allows us to regress the marginal vector field from \cref{eq:vectorfield} using only samples from $p_t\left(x \mid z\right)$ and $u_t\left(x \mid z \right)$.
Please refer to \citet{tongimproving} for a discussion on common forms for the conditional probability paths and vector fields.
\section{MPPCA Base Distributions}
\label{sect:mppca}
The initial probability distribution $p_0$ introduced in \cref{sect:cnfs} is typically a standard normal density, which allows for tractable density estimation using \cref{eq:pushforward} \cite{rezende2015variational, onken2021ot, lipmanflow}. 
\citet{tongimproving} showed how to conduct flow matching with samples from an arbitrary $p_0$; however, density estimation is no longer feasible if $p_0$ does not have an analytical likelihood expression.
In this work, we propose representing $p_0$ with MPPCA models, which we introduce in the following section.

\subsection{Low-Rank Mixture Models}
\begin{figure}
    \centering
    \includegraphics[width=\columnwidth]{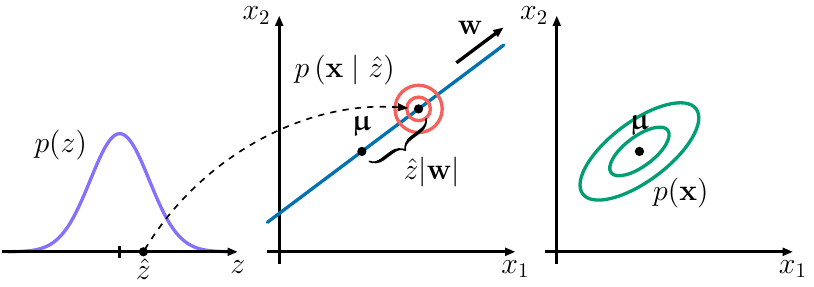}
    \caption{In the probabilistic PCA framework, samples from a low-dimensional latent space are projected to the data space by the factor loading matrix, which defines the principal component directions.}
    \label{fig:mppca}
\end{figure}
Principal component analysis (PCA) is a classic statistical technique for dimensionality reduction.
\citet{tipping1999mixtures} reformulate PCA within a maximum likelihood framework, resulting in an associated probability density.
Consider the following latent variable model:
\begin{align}
\label{eq:mppca}
\vect{x} &= \vect{W}\vect{z} + \vect{\mu} + \vect{\epsilon} \\
\vect{z} &\sim \mathcal{N}(\vect{0}, \vect{I}) \\
\vect{\epsilon} &\sim \mathcal{N}(\vect{0}, \sigma^2 \vect{I})
\end{align}
where $\vect{W}$ is a rectangular \textit{factor loading} matrix of size $d \times \ell$ and $\ell$ is a latent dimension such that $\ell \ll d$. 
The latent vector $\vect{z}$ is of length $\ell$, the mean vector $\vect{\mu}$ is of length $d$, and $\vect{\epsilon}$ is added noise with diagonal covariance $\sigma^2 \vect{I}$. 
For MPPCA, the noise is assumed to be isotropic; in the more general MFA framework, the noise assumption is relaxed to merely be diagonal.
In the case of isotropic noise, the implied conditional distribution is
\begin{equation}
    \label{eq:conditional}
    p\left(\vect{x} \mid \vect{z}\right) = \left( 2 \pi \sigma^2 \right)^{-d/2} \exp{\left\{ -\dfrac{1}{2\sigma^2} \lVert \vect{x} - \vect{W}\vect{z} - \vect{\mu}\rVert^2 \right\}}
\end{equation}
The Gaussian prior over the latent variables is given by
\begin{equation}
    \label{eq:prior}
    p\left(\vect{z}\right) = \left( 2 \pi \right)^{-\ell/2} \exp{\left\{ -\dfrac{1}{2} \vect{z}^\top \vect{z}\right\}}
\end{equation}
Thus, the marginal density over $\vect{x}$ can be expressed as
\begin{equation}
    \label{eq:marginal}
    p\left(\vect{x}\right) = \eta \lvert \vect{C} \rvert^{- 1 / 2} \exp{\left\{ -\dfrac{1}{2} \left(\vect{x} - \vect{\mu}\right)^\top \vect{C}^{-1} \left(\vect{x} - \vect{\mu}\right)\right\}}
\end{equation}
with normalizing constant $\eta = \left( 2 \pi \right)^{-d / 2}$ and model covariance $\vect{C} = \vect{W} \vect{W}^\top + \sigma^2 \vect{I}$.
The likelihood of a dataset $\{\vect{x}_0, \vect{x}_1, \ldots, \vect{x}_N\}$ is maximized when the columns of the factor loading matrix $\vect{W}$ span the principal subspace of the data \citep{tipping1999mixtures}.
\Cref{eq:marginal} defines the marginal likelihood for a single probabilistic principal component analyzer (PPCA); multiple PPCA models can be combined in a mixture to learn local principal axes and model more complex distributions \cite{tipping1999mixtures}. 
\Cref{fig:mppca} shows how a low-dimensional latent variable is projected to the data space for a single PPCA model, defining a Gaussian marginal distribution $p\left(\vect{x}\right)$.

\subsection{Fitting MPPCA Models}
Consider a mixture of $K$ probabilistic principal component analyzers as presented in \cref{eq:mppca}.
The log-likelihood of a dataset $\{\vect{x}_0, \vect{x}_1, \ldots, \vect{x}_N\}$ is
\begin{equation}
    \label{eq:loglike}
    \mathcal{L} = \sum_{i=1}^N \log \left\{ \sum_{k=1}^K \pi_k p\left(\vect{x}_n \mid k \right) \right\}
\end{equation}
where $p\left(\vect{x} \mid k \right)$ represents the density of the $k$th PPCA model and $\pi_k$ is the corresponding mixing proportion, with $\pi_k \geq 0$ and $\sum_{k=1}^K \pi_k = 1$.
\citet{tipping1999mixtures} derive an iterative expectation-maximization procedure with closed-form updates for parameters $\vect{W}_k$, $\vect{\mu}_k$, $\pi_k$, and $\sigma^2_k$ for components $k = 1, \ldots, K$ that is guaranteed to find a local maximum of \cref{eq:loglike}.
The EM procedure can also be performed in a memory-efficient fashion for large or high-dimensional datasets by accumulating sufficient statistics over minibatches \cite{richardson2018gans}.

\section{Experiments}
\label{sect:experiments}
This section first presents our datasets and evaluation metrics before discussing experimental results.

\subsection{Datasets}
We perform density estimation on three tabular datasets from the UC Irvine Machine Learning Repository, following the preprocessing steps of \citet{papamakarios2017masked}. 
These datasets are widely used in flow-based generative modeling benchmarks \cite{durkan2019neural,grathwohl2018ffjord,onken2021ot}.
Furthermore, we train flows on three datasets of natural images: FashionMNIST \citep{xiao2017fashion}, CelebA \citep{liu2015deep}, and CIFAR-10 \citep{krizhevsky2009learning}.
We crop, normalize, and resize the CelebA dataset to both $32 \times 32$ and $64 \times 64$.

\subsection{Metrics}
\label{sect:metrics}
We compute the following metrics to evaluate the quality of the learned generative models and quantify the impact of learning an MPPCA base distribution:
\begin{itemize}
    \item \textit{Average log-likelihood}: A held-out test dataset is transformed by integrating backwards through the learned map $\phi$ and the log-likelihood of each data point is computed according to \cref{eq:pushforward}. A \textit{higher} value indicates that the learned flow density more closely matches the samples.
    \looseness=-1\item \textit{Number of statistically different bins (NDB)}: The NDB metric is a simple method to evaluate generative models based on relative proportions of samples that fall into predetermined bins \citep{richardson2018gans}. 
    A \textit{lower} value indicates that the learned distribution more closely represents the data distribution.
    We divide by the number of bins $C$ and report the \textit{proportion} of statistically different bins.
    \item \textit{Fr{\'e}chet inception distance metric (FID)}: 
    FID is a standard metric for evaluating the quality of generated images commonly found in generative adversarial network literature \citep{heusel2017gans}. 
    In particular, we use the embeddings generated by the third layer of a pretrained InceptionV3 model, as is standard, to compute the FID metric. A \textit{lower} value indicates that the learned distribution more closely represents the data distribution.
    \item \textit{Number of function evaluations (NFE)}: We report the average number of function evaluations required by an adaptive-step ODE solver to produce the integration map $\phi$ at a prespecified numerical tolerance.
    A \textit{higher} NFE value indicates that the learned ODE dynamics are more complex, resulting in more computational burden at inference time due to the increased number of solver steps \cite{finlay2020train,lipmanflow}.
\end{itemize}

We report the log-likelihood metric for the UCI datasets and the NDB metric for the image datasets.
In all experiments, we report the average required NFE as well as the total training time.
The total training time includes the time required to fit the MPPCA base distributions.

\subsection{Experimental Setup}
We parameterize the vector field $u_t$ with a simple multilayer perception (MLP) neural network for the tabular density estimation tasks and with a U-net \cite{ronneberger2015u,nichol2021improved} for the image generation tasks.
The U-net weights are updated with an exponential moving average to smooth out noisy parameter updates.
We formulate $p_0$ as an MPPCA model for our proposed technique and benchmark against a standard choice of $p_0 = \mathcal{N}(\vect{0}, \vect{I})$.
Flows are trained via stochastic gradient descent using the variance-preserving (VP) flow matching objective \citep{albergo2023stochastic}.
Furthermore, to quantify the simplifying effect of the MPPCA base, we also ablate the flow matching objective by considering a minibatch optimal transport objective \citep{tongimproving}.
The minibatch approximations to the OT map have been shown to create simpler flows that are more stable to train.
For experiments where $p_0$ is an MPPCA model, we first fit the model on the training dataset before training the flow separately.
Hyperparameters are listed in \Cref{tab:hyperparameters}, where $\alpha_0$ is the initial learning rate.
MPPCA hyperparameter values are similar to the values used by \citet{richardson2018gans}.

\begin{table}[!tb]
    \centering
    \caption{Hyperparameters}
    \begin{tabular*}{\columnwidth}{@{\extracolsep{\fill}}lcrccr@{}}
        \toprule
        \ & $K$ & $\ell$ & EM iter &  $\alpha_0$ & batch \\ \midrule
        \hepmass{}      & \num{10}  & \num{4}   & \num{10} & \num{5e-4} & \num{256} \\
        \miniboone{}    & \num{30}  & \num{6}   & \num{50} & \num{5e-4} & \num{64} \\
        \bsds{}         & \num{30}  & \num{6}   & \num{10} & \num{5e-4} & \num{256} \\
        \fashion{}      & \num{50}  & \num{6}   & \num{10} & \num{2e-4} & \num{128} \\
        \cifar{}        & \num{50}  & \num{10}  & \num{10} & \num{2e-4} & \num{128} \\
        \celeba{} $(32\times32)$       & \num{50}  & \num{10}  & \num{10} & \num{2e-4} & \num{128} \\
        \celeba{} $(64\times64)$      & \num{64}  & \num{16}  & \num{16} & \num{2e-4} & \num{128} \\
        \bottomrule
    \end{tabular*}
    \label{tab:hyperparameters}
\end{table}

\subsection{Results}
\label{sect:results}

\begin{table*}[t]
\centering
\caption{Density estimation results on tabular UCI datasets.} 
\label{tab:tabular}
\small
    \begin{tabular*}{\textwidth}{@{\extracolsep{\fill}}
        c
        l
        S[table-format=3.1(2.1)]
        S[table-format=3.1(1.1)]
        S[table-format=3.2(1.2)]
        S[table-format=3.1(1.1)]
    @{}}
    \toprule
     \multirow{2}{*}{\vspace{-5pt}Dataset} & \multirow{2}{*}{\vspace{-5pt}Model} & \multicolumn{2}{c}{Training} & \multicolumn{2}{c}{Testing} \\
    \cmidrule(lr){3-4} \cmidrule(lr){5-6} 
     &   & {epochs ($\downarrow$)} & {time (min)  ($\downarrow$)} & {log-likelihood  ($\uparrow$)} & {NFE  ($\downarrow$)}\\
    \midrule
    \multirow{4}{*}{\shortstack[*]{\textbf{\hepmass{}}\\$d=21$}}  
        & VP-MPPCA  & \bfseries 34.6 \pm 5.9 & \bfseries 1.7 \pm 0.3 & -27.00 \pm 0.03           & \bfseries 50.0 \pm 0.2 \\
        & OT-MPPCA  & 48.6 \pm 7.2           & 8.5 \pm 1.2           & \bfseries -26.93 \pm 0.01 & 52.5 \pm 1.4           \\
        & VP-Normal & 40.8 \pm 5.7           & 1.8 \pm 0.2           & -27.95 \pm 0.01           & 61.2 \pm 0.6           \\
        & OT-Normal & 52.6 \pm 6.2           & 9.1 \pm 1.0           & -27.91 \pm 0.01           & 63.2 \pm 0.7           \\

    \midrule                    
    \multirow{4}{*}{\shortstack[*]{\textbf{\miniboone{}}\\$d=43$}}  
        & VP-MPPCA  & 24.4 \pm 7.1           & \bfseries 0.3 \pm 0.1 & \bfseries -30.97 \pm 0.46 & \bfseries 43.1 \pm 1.0 \\
        & OT-MPPCA  & \bfseries 21.0 \pm 7.2 & 0.4 \pm 0.1           & -31.42 \pm 0.45           & 43.5 \pm 0.7           \\
        & VP-Normal & 53.6 \pm 14.2          & 0.5 \pm 0.1           & -50.37 \pm 0.99           & 57.0 \pm 1.0           \\
        & OT-Normal & 82.2 \pm 12.0          & 1.1 \pm 0.2           & -48.40 \pm 0.55           & 62.4 \pm 1.4           \\
    \midrule
    \multirow{4}{*}{\shortstack[*]{\textbf{\bsds{}}\\$d=63$}} 
        & VP-MPPCA  & \bfseries 31.6 \pm 7.1 & \bfseries 10.6 \pm 1.0 & 126.64 \pm 1.68           & 50.9 \pm 1.1           \\
        & OT-MPPCA  & 32.8 \pm 5.6           & 25.0 \pm 3.2           & \bfseries 127.25 \pm 1.11 & \bfseries 50.7 \pm 0.6 \\
        & VP-Normal & 140.6 \pm 21.0         & 17.2 \pm 2.7           & 69.76 \pm 1.67            & 102.8 \pm 0.5          \\
        & OT-Normal & 171.0 \pm 15.6         & 102.4 \pm 9.3          & 86.14 \pm 1.39            & 113.8 \pm 0.5          \\
    \bottomrule
    \end{tabular*}
\end{table*}%

\begin{table*}[t]
\centering
\caption{Generative modeling results on image datasets.}
\label{tab:images}
\small
    \begin{tabular*}{\textwidth}{@{\extracolsep{\fill}}
        c
        l
        r
        S[table-format=3.1(1.1)]
        S[table-format=2.2(1.2)]
        S[table-format=3.1(1.1)]
        S[table-format=3.1(1.1)]
    @{}}
    \toprule
     \multirow{2}{*}{\vspace{-5pt}Dataset} & \multirow{2}{*}{\vspace{-5pt}Model} & \multicolumn{2}{c}{Training} & \multicolumn{3}{c}{Testing} \\
    \cmidrule(lr){3-4} \cmidrule(lr){5-7}
     &   & {epochs} & {time (min)} & {NDB/$C$  ($\downarrow$)} & {NFE ($\downarrow$)} & {FID ($\downarrow$)}\\
    \midrule
    \multirow{4}{*}{\shortstack[*]{\textbf{\fashion{}}\\$[28 \times 28 \times 1]$}}
        & VP-MPPCA  & 100   & 172.3 \pm 6.0 & 0.52 \pm 0.03 & 32.8 \pm 0.3 & 131.7 \pm 2.2 \\
        & OT-MPPCA  & 100   & 187.5 \pm 7.9 & \bfseries 0.37 \pm 0.02 & \bfseries 26.0 \pm 0.0 & \bfseries 92.5 \pm 4.9 \\
        & VP-Normal  & 100   & 299.8 \pm 1.9 & 0.92 \pm 0.01 & 38.2 \pm 0.2 & 133.3 \pm 3.3 \\
        & OT-Normal  & 100   & 183.9 \pm 7.6 & 0.89 \pm 0.01 & 38.0 \pm 0.0 & 136.2 \pm 7.6 \\
    \midrule
    \multirow{4}{*}{\shortstack[*]{\textbf{\cifar{}}\\$[32 \times 32 \times 3]$}}
        & VP-MPPCA  & 100   & 224.9 \pm 2.2 & 0.33 \pm 0.00 & \bfseries 26.0 \pm 0.0 & 289.9 \pm 2.5 \\
        & OT-MPPCA  & 100   & 172.2 \pm 1.4 & \bfseries 0.30 \pm 0.02 & \bfseries 26.0 \pm 0.0 & \bfseries 268.4 \pm 2.4 \\
        & VP-Normal  & 100   & 246.5 \pm 1.4 & 0.88 \pm 0.00 & 38.0 \pm 0.0 & 303.2 \pm 8.5 \\
        & OT-Normal  & 100   & 199.1 \pm 2.6 & 0.84 \pm 0.00 & 32.0 \pm 0.0 & 305.9 \pm 3.4 \\
    \midrule
    \multirow{4}{*}{\shortstack[*]{\textbf{\celeba{}}\\$[32 \times 32 \times 3]$}}
        & VP-MPPCA  & 50   & 319.6 \pm 1.3 & 0.44 \pm 0.03 & 26.1 \pm 0.1 & 304.7 \pm 1.2 \\
        & OT-MPPCA  & 50   & 501.5 \pm 1.8 & \bfseries 0.32 \pm 0.02 & \bfseries 26.0 \pm 0.0 & \bfseries 237.1 \pm 3.8 \\
        & VP-Normal  & 50   & 327.2 \pm 1.5 & 0.83 \pm 0.01 & 50.2 \pm 4.6 & 289.9 \pm 2.4 \\
        & OT-Normal  & 50   & 495.0 \pm 1.8 & 0.86 \pm 0.01 & 37.9 \pm 0.1 & 290.9 \pm 2.3 \\
    \midrule
    \multirow{4}{*}{\shortstack[*]{\textbf{\celeba{}}\\$[64 \times 64 \times 3]$}}
        & VP-MPPCA  & 50   & 451.3 \pm 0.0 & 0.68 \pm 0.00 & 26.7 \pm 0.0 & 285.3 \pm 0.0 \\
        & OT-MPPCA  & 50   & 351.3 \pm 0.0 & \bfseries 0.38 \pm 0.00 & \bfseries 26.0 \pm 0.0 & \bfseries 239.6 \pm 0.0 \\
        & VP-Normal  & 50   & 540.6 \pm 2.4 & 0.89 \pm 0.02 & 56.0 \pm 0.0 & 387.9 \pm 7.5 \\
        & OT-Normal  & 50   & 407.6 \pm 1.1 & 0.89 \pm 0.01 & 44.0 \pm 0.0 & 377.5 \pm 4.2 \\
    \bottomrule
    \end{tabular*}
\end{table*}%

\begin{figure*}[htbp]
    \centering
    \begin{subfigure}{\textwidth}
        \begin{subfigure}{0.162\textwidth}
            \centering
            \includegraphics[width=\linewidth]{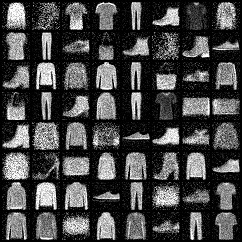}
        \end{subfigure}
        \begin{subfigure}{0.162\textwidth}
            \centering
            \includegraphics[width=\linewidth]{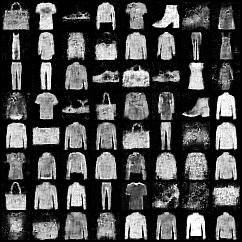}
        \end{subfigure}
        \begin{subfigure}{0.162\textwidth}
            \centering
            \includegraphics[width=\linewidth]{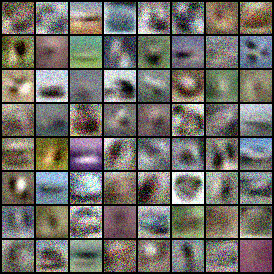}
        \end{subfigure}
        \begin{subfigure}{0.162\textwidth}
            \centering
            \includegraphics[width=\linewidth]{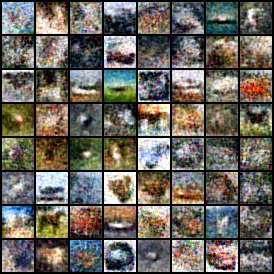}
        \end{subfigure}
        \begin{subfigure}{0.162\textwidth}
            \centering
            \includegraphics[width=\linewidth]{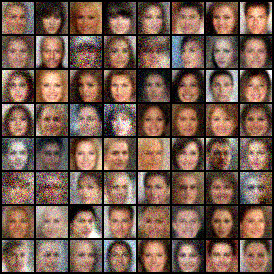}
        \end{subfigure}
        \begin{subfigure}{0.162\textwidth}
            \centering
            \includegraphics[width=\linewidth]{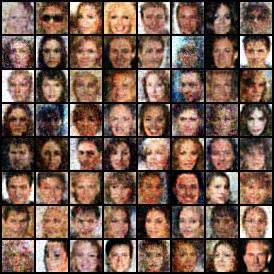}
        \end{subfigure}
        \caption{Samples from OT-MPPCA.}
        \label{fig:otmppca}
        \vspace{1em} 
    \end{subfigure}
    \begin{subfigure}{\textwidth}
        \begin{subfigure}{0.162\textwidth}
            \centering
            \includegraphics[width=\linewidth]{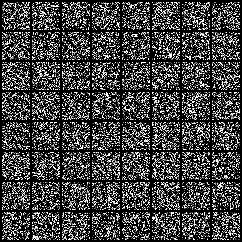}
        \end{subfigure}
        \begin{subfigure}{0.162\textwidth}
            \centering
            \includegraphics[width=\linewidth]{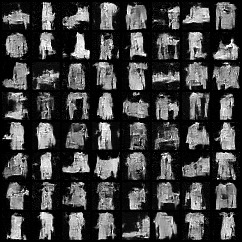}
        \end{subfigure}
        \begin{subfigure}{0.162\textwidth}
            \centering
            \includegraphics[width=\linewidth]{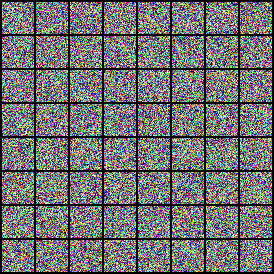}
        \end{subfigure}
        \begin{subfigure}{0.162\textwidth}
            \centering
            \includegraphics[width=\linewidth]{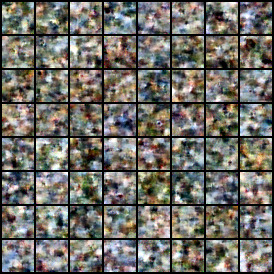}
        \end{subfigure}
        \begin{subfigure}{0.162\textwidth}
            \centering
            \includegraphics[width=\linewidth]{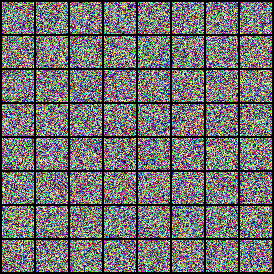}
        \end{subfigure}
        \begin{subfigure}{0.162\textwidth}
            \centering
            \includegraphics[width=\linewidth]{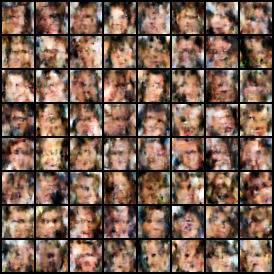}
        \end{subfigure}
        \caption{Samples from OT-Normal.}
        \label{fig:otnormal}
    \end{subfigure}
    \caption{For each image dataset, sample pairs show samples from the learned model after the first (left) and last (right) training epoch.}
    \label{fig:samples}
\end{figure*}

We implement early stopping for the density estimation tasks, stopping training if there is no improvement in validation loss for $10$ epochs.
For the image generation tasks, we train for a fixed number of epochs across all trials, hypothesizing that the warm-start provided by the MPPCA base density will lead to improved performance given a constrained training budget.
We sweep across five random seeds for the density estimation experiments and three random seeds for the image generation experiments.
Code to reproduce the experimental results is available at \texttt{\url{https://github.com/sisl/LowRankLatentGMMs}}.

\Cref{tab:tabular} shows the results for density estimation on the UCI datasets.
Early stopping generally triggers earliest when training flows with an MPPCA base, as the average number of required training epochs decreases dramatically for VP-MPPCA and OT-MPPCA models on the MINIBOONE and BSDS300 datasets.
The computational savings obtained from early stopping outweigh the additional overhead cost of fitting the more expressive base distribution, as evidenced by the relatively short overall training times for models with an MPPCA base.
Flows with MPPCA base densities obtain the highest average log-likelihood values on all three datasets.
Interestingly, VP-MPPCA achieves the lowest average NFE on two of the three datasets, implying that the simplifying effect of learning an expressive base distribution does not necessarily benefit from the additional application of optimal-transport-based inductive biases.

\Cref{tab:images} shows the results for the image generation experiments.
Once again, flows with MPPCA base densities require the fewest average NFEs at inference time, implying that VP-MPPCA and OT-MPPCA learn simpler integration paths.
Furthermore, the OT-MPPCA architecture achieves the lowest NDB/$C$ and FID scores across all image datasets given a fixed number of training epochs, which indicates that learning an expressive base density helps the model generate samples that are a closer match to the ground-truth images.
Learning the MPPCA base density warm-starts the training procedure---a phenomenon clearly visible in \cref{fig:samples}.
Each row shows image pairs of generated samples from the FashionMNIST, CIFAR-10, and CelebA datasets.
The leftmost image in each pair shows samples generated after the first epoch, while the rightmost image shows samples generated after the final epoch.
By initializing the normalizing flow with an MPPCA base density, the model starts with a structured prior that closely approximates the data distribution. 
This warm-start significantly reduces the burden on the flow to learn complex transformations, allowing it to refine rather than entirely reshape the base distribution. 
As a result, even after a single training epoch, generated samples already exhibit meaningful resemblance to ground truth images. 

\section{MPPCA Considerations}
\label{sect:considerations}
We next examine computational considerations and the importance of the latent factors hyperparameter.

\subsection{Computational Overhead}
\label{sect:computation}
Learning an MPPCA base introduces additional computational overhead due to the need to fit the mixture model separately before flow training. 
Unlike a standard normal base, MPPCA requires learning cluster means, factor loading matrices, noise variances, and mixing coefficients.
The total number of parameters for an MPPCA model with $K$ components is $K(d\ell + d + 1) +(K-1)$.
The time required to fit the MPPCA model is minimal compared to the overall training process, accounting for less than $3\%$ of the total training time across all image generation experiments, as shown in \cref{tab:fitting-times}.

\begin{table}[t]
\centering
\caption{Time required to fit MPPCA via EM versus total training times.} 
\label{tab:fitting-times}
\small
    \begin{tabular*}{\columnwidth}{@{\extracolsep{\fill}}
        c
        l
        S[table-format=1.1(1.1)]
        S[table-format=3.1(1.1)]
    @{}}
    \toprule
     {Dataset} & {Model} & {EM time (min)} & {Total (min)}\\
    \midrule
    \multirow{2}{*}{\shortstack[*]{\textbf{\fashion{}}}}  
        & VP-MPPCA  & 0.3 \pm 0.0     & 93.5 \pm 0.1  \\
        & OT-MPPCA  & 0.3 \pm 0.0     & 94.3 \pm 0.3 \\
    \midrule                    
    \multirow{2}{*}{\shortstack[*]{\textbf{\cifar{}}}}  
        & VP-MPPCA  & 0.7 \pm 0.0     & 85.3 \pm 0.0  \\
        & OT-MPPCA  & 0.7 \pm 0.0     & 86.3 \pm 0.0  \\
    \midrule
    \multirow{2}{*}{\shortstack[*]{\textbf{\celeba{}}\\$[32 \times 32 \times 3]$}} 
        & VP-MPPCA  & 4.7 \pm 0.1     & 192.3 \pm 0.1  \\
        & OT-MPPCA  & 4.7 \pm 0.0     & 194.6 \pm 0.5  \\
    \bottomrule
    \end{tabular*}
\end{table}%

\subsection{The Effect of Latent Factors on Model Performance}

\begin{figure}[t]
    \begin{subfigure}[t]{\columnwidth}
    \centering
        \includegraphics[width=\linewidth]{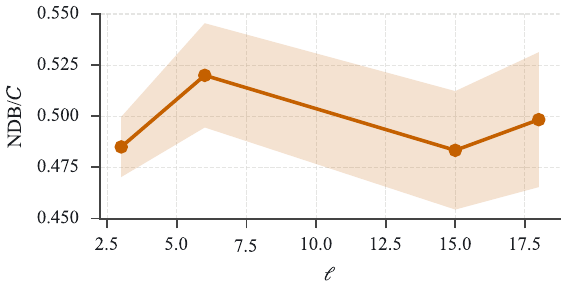}
        \label{fig:ndbs}
        \vspace{-1.5em}
        \caption{NDB/$C$ versus number of latent factors}
    \end{subfigure}
    \begin{subfigure}[t]{\columnwidth}
    \centering
        \vspace{1em}
        \includegraphics[width=\linewidth]{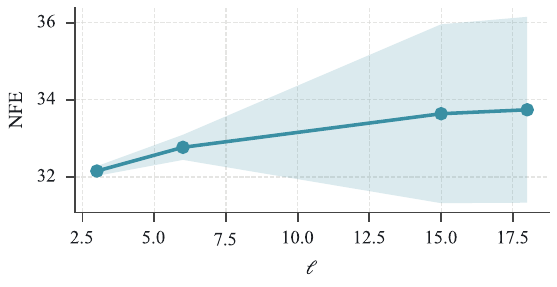}
        \label{fig:nfes}
        \vspace{-1.5em}
        \caption{NFEs versus number of latent factors}
    \end{subfigure}
    
    \hfill
\caption{Performance metrics on the FashionMNIST dataset as a function of latent factors.}
\label{fig:factors}
\end{figure}

We consider the FashionMNIST dataset and sweep over a range of latent factor counts for the latent MPPCA distribution, recording the average NDB/$C$ and NFE scores across test batches for three random seeds.
The ranges of performance metrics are visualized in \cref{fig:factors}.
Overall, performance is relatively steady across the sweep of latent factors, indicating that MPPCA models with even a small number of latent factors can provide a meaningful inductive bias for the learning process.
\citet{richardson2018gans} showed that MPPCA models with $10$ or fewer latent factors can adequately approximate high-dimensional distributions, which is consistent with \cref{fig:factors}.

\section{Discussion}
\label{sect:conclusion}
We present a technique to simplify flow matching transformations using low-rank mixture base distributions.
Mixtures of probabilistic principal component analyzers are a parameter-efficient model that can be updated analytically via expectation-maximization, even in high-dimensional spaces.
Initializing the flow matching with an MPPCA base density serves as an inductive bias on the training process, reducing the burden on the flow to learn complex transformations.
We show that MPPCA base distributions can be combined with continuous normalizing flows to generate higher-quality samples and reduce the number of required integration steps at inference time compared to standard normal base densities.

Future work will explore techniques to scale the MPPCA fitting procedure to even higher dimensions, such as feature sampling or initializing mixture component clusters with constrained K-means.
We will also evaluate the simplifying effect of MPPCA base densities when combined with other flow matching objectives such as action matching \citep{neklyudov2023action} and Schr{\"o}dinger Bridge flow matching \citep{tongimproving}.

\section*{Acknowledgments}
Toyota Research Institute (TRI) provided funds to assist the authors with their research, but this article solely reflects the opinions and conclusions of its authors and not TRI or any other Toyota entity.

\renewcommand*{\bibfont}{\footnotesize}
\printbibliography

\end{document}